\documentclass{article}
\usepackage{PRIMEarxiv}
\usepackage{amsmath,amsfonts}
\usepackage{algorithmic}
\usepackage{algorithm}
\usepackage{array}
\usepackage{textcomp}
\usepackage{stfloats}
\usepackage{url}
\usepackage{verbatim}
\usepackage{graphicx}
\usepackage{subcaption}
\usepackage{cite}
\usepackage{booktabs}
\usepackage{graphicx}
\usepackage{subcaption}

\usepackage[table]{xcolor}  
 \usepackage{booktabs} 
 \usepackage{xcolor}
 \usepackage{multirow} 
\usepackage[utf8]{inputenc} 
\usepackage[T1]{fontenc}    
\usepackage{hyperref}       
\usepackage{cleveref}
\usepackage{url}            
\usepackage{booktabs}       
\usepackage{amsfonts}       
\usepackage{nicefrac}       
\usepackage{microtype}      
\usepackage{lipsum}
\usepackage{fancyhdr}       
\usepackage{graphicx}       
\graphicspath{{media/}}     

\title{LayersReg: A Layer-by-Layer Progressive Regressor for Reliable Intraoperative 3D/2D Registration
\thanks{*Chuan Chen and Feng Yin are the corresponding authors.} 
}

\author{
  Xiyuan Wang, Zhenchao Wang, Xinran Chen \\
  School of Biological Science and Medical Engineering\\
Southeast University \\
  Nanjing\\
  \texttt{\{230249082, 230238672, 213222326\} @seu.edu.cn} \\
   \And
  Junkai Liu \\
  School of Engineering, College of Engineering and Physical Sciences \\
  University of Birmingham\\
  Birmingham\\
  \texttt{jxl1920@student.bham.ac.uk} \\
    \And
   Chuan Chen* \\
  School of Biological Science and Medical Engineering\\
  Southeast University\\
  Nanjing\\
  \texttt{chuanchen@seu.edu.cn} \\
  \And
   Feng Yin* \\
  State Key Laboratory of Digital Medical Engineering \\
  Southeast University\\
  Nanjing\\
  \texttt{yinfeng@seu.edu.cn} \\
}

\begin{document}
\maketitle

\begin{abstract}
3D/2D registration serves as a cornerstone technique in surgical navigation. Traditional iterative optimization algorithms suffer from low efficiency and high failure rates in intraoperative settings. Deep learning-based methods reformulate registration from iterative optimization to a regression problem that maps image appearance features to spatial pose, typically achieving improved real-time performance and accuracy. However, such learnable methods are confined to memory-driven retrieval of specific pose features rather than understanding the task of image alignment itself, which limits their generalization in complex scenarios. We propose LayersReg, a pioneering regression paradigm that endows the model with 3D anatomical awareness and searches for the correct pose in a progressive, layer-by-layer manner. Inspired by the iterative pose-searching optimization criterion of classical registration, LayersReg searches for correlations between the moving and fixed images in feature space, capturing the trend of pixel flow and thereby converging iteratively toward the correct spatial pose transformation. We further design a coupling of node-wise regression with the progressive registration framework to enhance the model's perception of spatial pose changes. Experimental results demonstrate that under large offsets and multimodality conditions, LayersReg achieves high accuracy on both X-ray/CT registration (0.68°, 1.41 mm) and slice localization (0.73°, 1.55 mm) tasks, outperforming existing state-of-the-art methods while meeting the intraoperative demands for precision and real-time capability.

\end{abstract}


\section{Introduction}
\begin{figure*}
    \centering
    \includegraphics[width=0.98\textwidth]{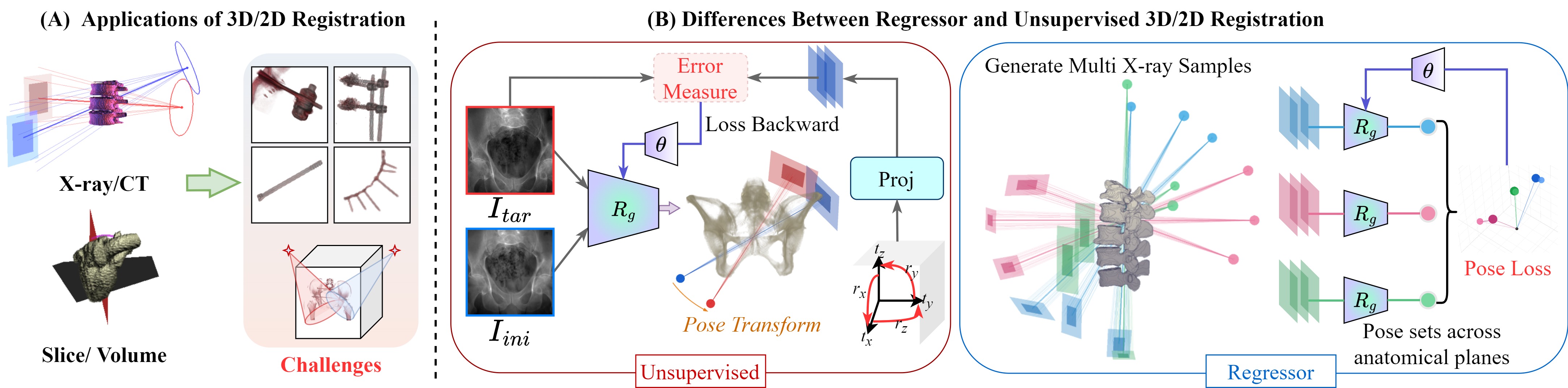}
   \caption{(A) Application scenarios of 3D/2D registration and existing challenges (e.g., occlusion, large misalignment). (B) Comparison of two training pipelines, unsupervised registration (red) and regressor-based (blue). $R_g$ is the registration model.}
    \label{f1}
\end{figure*}

3D/2D registration is a core technique in modern surgical navigation, establishing the coordinate transformation between the patient's physical space and the pre-operative image space \cite{in1,in2}. By spatially aligning intra-operative 2D images (e.g., X-ray, 2D ultrasound) with pre-operative 3D volumetric data (e.g., MRI, CT), this technique provides real-time localization and guidance for surgical instrument placement. In practice, 3D/2D registration covers two tasks which are X-ray/CT and slice-to-volume (S2V) registration, as illustrated in \cref{f1}(A). Currently, iterative optimization methods based on image similarity metrics \cite{iter1, NGI, OPT-GO} dominate the field due to their high accuracy. However, these methods are prone to local minima, especially under surgical instrument occlusion or large initial misalignment, which can cause severe efficiency degradation or even complete failure. In clinical practice, this often means relying on experts to manually annotate anatomical landmarks for initialization, a time-consuming and laborious process that hinders real-time surgical navigation.

In recent years, pose regression \cite{miao1,miao3} has become the dominant paradigm in 2D/3D registration, delivering real-time prediction with greater alignment efficiency and capture range than traditional iterative optimization. These methods employ CNNs to extract features from X-ray images and learn a direct nonlinear mapping to 6-degree-of-freedom (6DoF) rigid poses. However, such regressors exhibit limited generalization in real surgical scenarios (\cref{f1}(A)), owing to two fundamental issues. First, they generally adopt supervised (or synthetically self-supervised) methods. As shown in \cref{f1}\textcolor{blue}{(B)}, training relies on large-scale synthetic digitally reconstructed radiographs (DRRs) to model the distribution of image appearance conditioned on annotated poses, thereby indirectly learning the 3D structure of CT to infer X-ray pose. By contrast, unsupervised or differentiable registration methods adjust misalignment parameters via gradient flow, driven by the error between registered images (\cref{f1}\textcolor{red}{(B)}). In essence, existing regressors leverage the memorization of appearance-pose mapping, rather than learning to perceive and correct misalignment through gradient-based optimization as unsupervised methods. Second, the majority of regressors rely solely on a single X-ray image, without access to 3D anatomical priors. Projection discards depth and occlusion information, preventing the network from perceiving discrepancies between the predicted pose and the true anatomy. Later works \cite{PRSCS, DVAP-Reg} have introduced back-projection to reconstruct intraoperative pose CT from X-rays. Yet accurate reconstruction from a single X-ray remains an open challenge, typically requiring orthogonal views to recover missing 3D information, a condition seldom met in practice.

In this paper, we propose LayersReg, designed to overcome the aforementioned limitations of existing pose regressors. Specifically, to address the lack of 3D anatomical priors, we extract features from the CT volume, enabling cross-dimensional interaction between 2D X-ray and 3D volumetric features. This brings richer spatial depth awareness into pose prediction and substantially expands the capture range. To address the failure to genuinely learn registration via error minimization, LayersReg abandons the conventional one-shot static mapping, and instead reformulates pose prediction as a progressive internal refinement of the predicted pose. Following the core principle of iterative registration, the method starts from an initial misalignment, iteratively measures the feature discrepancy between the moving and fixed images, and leverages this error signal to drive the pose estimate progressively toward the correct relative transformation between the intraoperative 2D image and the preoperative 3D volume.

The workflow of LayersReg is divided into feature extraction followed by pose regression. Firstly, we address the brittleness of previous regressors under surgical instrument occlusion. Trained solely with pose labels, they tend to mine regions statistically correlated with pose, a dependency that easily breaks down under occlusion. LayersReg employs hybrid autoencoders to extract compact latent representations from intraoperative images and preoperative volumes, preserving structural integrity while discarding appearance-level noise. We further inject compact 3D geometric cues into the 2D feature stream, achieving adaptive compensation for missing depth information. Secondly, following the energy minimization principle of registration, we construct a cross-correlation volume between the dual-stream features and capture the alignment trend from feature residuals. The regression backbone takes this correlation and trend as dual driving signals, and realizes deep state interaction via a designed Mamba module to aggregate pose cues. Furthermore, we introduce a node-wise regression strategy that treats the pose output at each stage as an intermediate node converging to the final pose, improving both accuracy and robustness in large-displacement scenarios. We extensively validate LayersReg across seven datasets spanning multiple modalities, anatomical sites, and surgical scenarios, including real X-ray, US volume localization, and multimodal settings, consistently demonstrating high accuracy and strong generalization. Our contributions are threefold:
\begin{enumerate}
    \item We propose LayersReg, a Mamba-based regression paradigm that embeds iterative refinement into pose prediction via closed-loop feature residual feedback, achieving 3D-aware, registration-driven pose estimation with deep 6-DoF coupling via node-wise regression.

\item We introduce hybrid autoencoders for robust feature extraction and dynamic 3D-to-2D fusion, injecting anatomical depth cues while preventing overfitting to pose-appearance patterns.

\item LayersReg demonstrates reliable and accurate performance across diverse datasets and surgical scenarios, establishing itself as an effective solution for intraoperative 3D/2D registration.
\end{enumerate}

\section{Related Work}
\subsection{Regression-based 3D/2D Registration}
Pose regression models have emerged as a key focus in recent rigid registration studies for pose estimation. To address adaptability challenges, some methods \cite{reg2,reg3,reg4,reg5,reg6} enhance performance through pose-supervised pre-training of regressors followed by secondary fine-tuning. This tuning centers on minimizing errors between registered and target images, employing self- or weakly-supervised \cite{reg1,reg7} optimization to refine predictions and bolster model generalization. The process can weigh losses in expert-annotated RoI regions, driving the model to select effective areas and resist interference. However, these methods face two major challenges: first, they require iterative optimization \cite{reg8,reg12,reg13,reg14}, along with registration-specific image error metrics \cite{OPT-GO}. Otherwise, accuracy will remain insufficient. Second, they heavily depend on expert-labeled intraoperative images and preoperative volumes, making large-scale training impractical despite preoperative planning involving labeled segmentation.

To tackle large offsets between target and initial images, some approaches \cite{back7,back8,back1,back4} use CNNs to back-project 2D X-ray images into 3D space. This method abandons planar pose estimation in favor of 3D/3D rigid registration, comparing reconstructed volumes with initial ones to predict rigid transformations (3D regression networks). Back-projection substantially enriches depth information, enabling better adaptation to large offsets and intuitive pose prediction. Yet, limitations remain evident: first, 3D regressor performance relies on reconstructed volume quality, and while studies \cite{back2,back3} have advanced high-quality volume generation, additional views are still needed to supplement 3D structures; second, compared to 2D regressors, 3D feature capture incurs higher overhead, and the approach is unsuitable for slice/volume registration due to slices lacking complete 3D information, impeding effective reconstruction.
\subsection{Geometric-based 3D/2D Registration}
Beyond regression tasks, neural networks are also employed to annotate anatomical landmarks in intraoperative/preoperative images \cite{gem1,gem8}, thereby enhancing the efficiency and accuracy of iterative optimization. Some methods train CNNs to learn anatomical structures in intraoperative images of patients under specific poses and predict keypoints or planar segmentation masks \cite{gem4,gem6,gem8}. In the registration pipeline, these approaches utilize fitted CNNs to extract anatomical structures from initial pose images and target images, mapping them into a 3D volume to ensure correct positioning. After obtaining the 2D-3D mapping, an optimizer rapidly solves for the correct pose, achieving exact alignment \cite{gem7}. This method not only improves accuracy but also applies to both X-ray/CT \& S2V scenarios, offering greater generality. However, annotation is time-consuming and must be tailored to specific anatomical sites. Training a unified annotation network for heterogeneous multimodal tasks, integrating different modalities simultaneously, remains highly challenging.
\begin{figure*}
    \centering
    \includegraphics[width=1\textwidth]{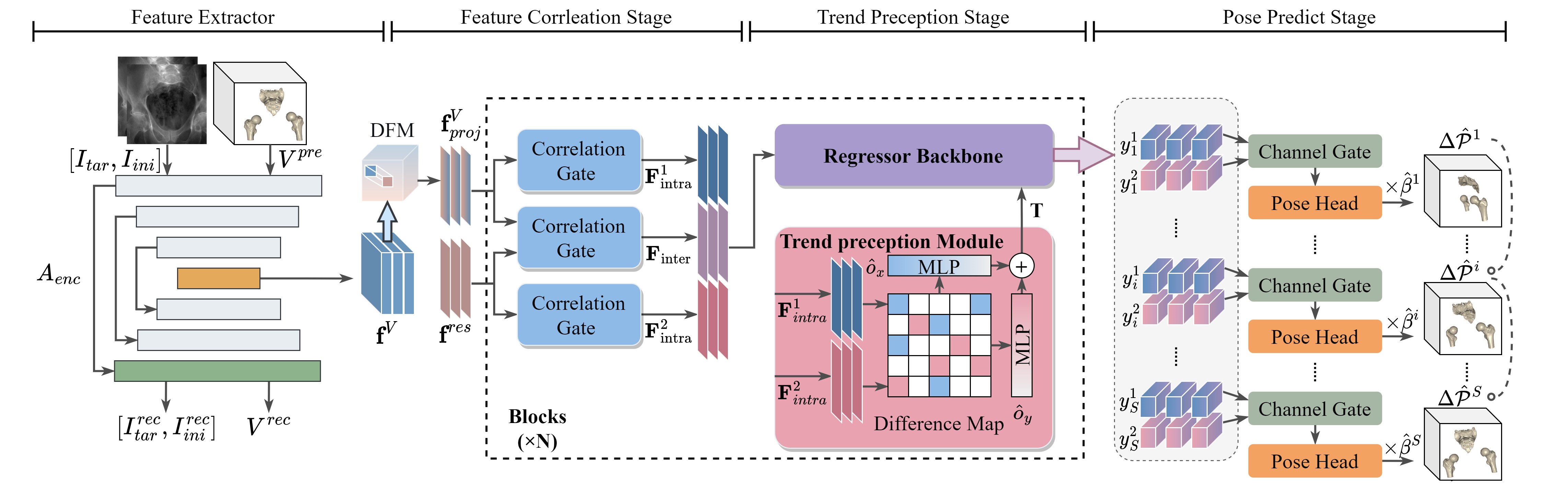}
    \caption{Implementation flowchart of LayersReg. Latent features are extracted by 2D and 3D encoders ($A^{2d}_{enc}$, $A^{3d}_{enc}$). Depth Feature Modulation (DFM) is used to reduce dimensionality and fuse 3D depth cues. The regressor outputs \(y^1_i\) and \(y^2_i\) at each internal stage and predicts the pose via channel‑wise gating. A learnable scalar \(\beta\) controls the step size by which each node advances toward the final pose. We adopt \(\mathcal{L}_{pos}\) for deep supervision of the pose. We employ \(\mathcal{L}^{2d}_{rec}\) and \(\mathcal{L}^{3d}_{rec}\) as self‑supervised reconstruction losses to train \(A^{2d}_{enc}\) and \(A^{3d}_{enc}\).
 }
    \label{f2}
\end{figure*}
\section{Methods}
An overview of our registration network is shown in \cref{f2}. It consists of 3D and 2D autoencoders ($A^{2d}_{enc}, A^{3d}_{enc}$) for stable feature extraction, and a core pose regression module comprising three components: dual correlation gating (DCG), a trend perception module (TPM), and a mamba-based backbone. In the following, we detail each module and review the principles of 3D/2D registration.

\subsection{Principles of 3D/2D Registration}
Given the preoperative volume \(V_{pre} \in \mathbb{R}^{h \times w \times d}\) and the target intraoperative image \(I_{tar} \in \mathbb{R}^{h \times w}\), optimization‑based registration methods aim to estimate the optimal relative pose transformation \(\Delta \mathcal{P}_{opt}\) by maximizing the similarity \(\mathcal{S}\) between \(I_{tar}\) and the 2D image rendered from \(V_{pre}\) after applying a candidate relative transformation \(\Delta \mathcal{P}\) on top of the initial pose \(\mathcal{P}_0\). This process can be formulated as:
\begin{equation}
    \Delta\mathcal{P}_{opt}
=
\arg\max_{\Delta \mathcal{P} \in SE(3)}
\mathcal{S}\left(\mathcal{U}(V_{pre}, \Delta \mathcal{P} \circ \mathcal{P}_0), I_{tar}\right).
\end{equation}
where $  \mathcal{U}(\cdot)  $ is the differentiable 3D-to-2D forward operator (i.e., DRR projection for X-ray/CT registration and plane slicing for S2V registration).

The regressor-based 3D/2D registration directly predicts the relative transformation. Given a known initial pose $\mathcal{P}_0$, the initial rendered image is computed as $  I_{ini} = \mathcal{U}(V_{pre}, \mathcal{P}_0)  $. The regressor $  R_\Theta  $ takes features extracted from the pair $  [I_{tar}, I_{ini}]  $ and outputs the predicted relative transformation $  \Delta \hat{\mathcal{P}}$. The training method of the regressor is as follows:
\begin{equation}
    \Theta^* = \arg\min_{\Theta} \, \delta \Bigl(   \Delta \hat{\mathcal{P}} = R_\Theta \bigl( f([I_{tar}, I_{ini}]) \bigr), \Delta \mathcal{P}_{gt} \Bigr).
    \label{eq2}
\end{equation}
 where \(f\) denotes feature extraction, \([\cdot,\cdot]\) denotes channel‑wise concatenation. $\Theta$ is model's weight, \(\Delta \mathcal{P}_{gt}\) represents the ground-truth pose, and \(\delta\) is the defined pose loss. LayersReg adopts the regression formulation of \cref{eq2}, yet enhances it by injecting $V_{pre}$ depth cues and progressively optimizing internal feature correspondences (via DCG and TPM) to bridge residual features and 6‑DoF poses.
 
\subsection{Hybrid Autoencoders for Feature Extraction}
The architectures of $A^{2d}_{enc}$ and $A^{3d}_{enc}$ are structurally similar, with minor differences in internal components. Their workflows are:
\begin{align}
    \bigl( \mathbf{f}^{res},\; [I^{rec}_{tar}, I^{\mathrm{rec}}_{ini}] \bigr) &= A^{2d}_{enc}([I_{tar}, I_{ini}]), \label{eq:2d} \\[4pt]
    \bigl( \mathbf{f}^{v},\; V^{rec} \bigr) &= A^{3d}_{enc}(V_{pre}). \label{eq:3d}
\end{align}

$\mathbf{f}^{res}, \mathbf{f}^{v}$ are the extracted features of image residual and volume. $I^{rec}_{tar},I^{rec}_{ini},V^{rec}$ are self-supervised reconstruction.
We adopt a modern network architecture, comprising a Dense Downsampling Layer (DDL), a refine layer, and a Dense Upsampling Layer (DUL). The DDL provides initial contextual priors and expands the channel receptive field: given an input size of $(H, W)$, it downsamples to ($\frac{H}{16}, \frac{W}{16}$). The DDL comprises four layers of basic (3D/2D) convolutions (kernel size = 3, stride = 2), followed by group normalization (GN) and SELU activation.

The refine layer enhances fundamental features: in $A^{2d}_{enc}$, Vision Mamba's (ViM) \cite{Vim} bi-SSM blocks are employed to boost long-sequence modeling, adaptively extracting key information and remove noise. In $A^{3d}_{enc}$, a combination of 3D depthwise convolution (DwConv) and 3D feed-forward network (FFN)  is used, followed by 3D SEblock \cite{SE}. The DUL restores the input image, similar to DDL; it additionally incorporates skip connections and transposed convolutions for feature fusion and resolution recovery. 
\subsection{Depth Feature Modulation}
\begin{figure}
    \centering
    \includegraphics[width=0.46\textwidth]{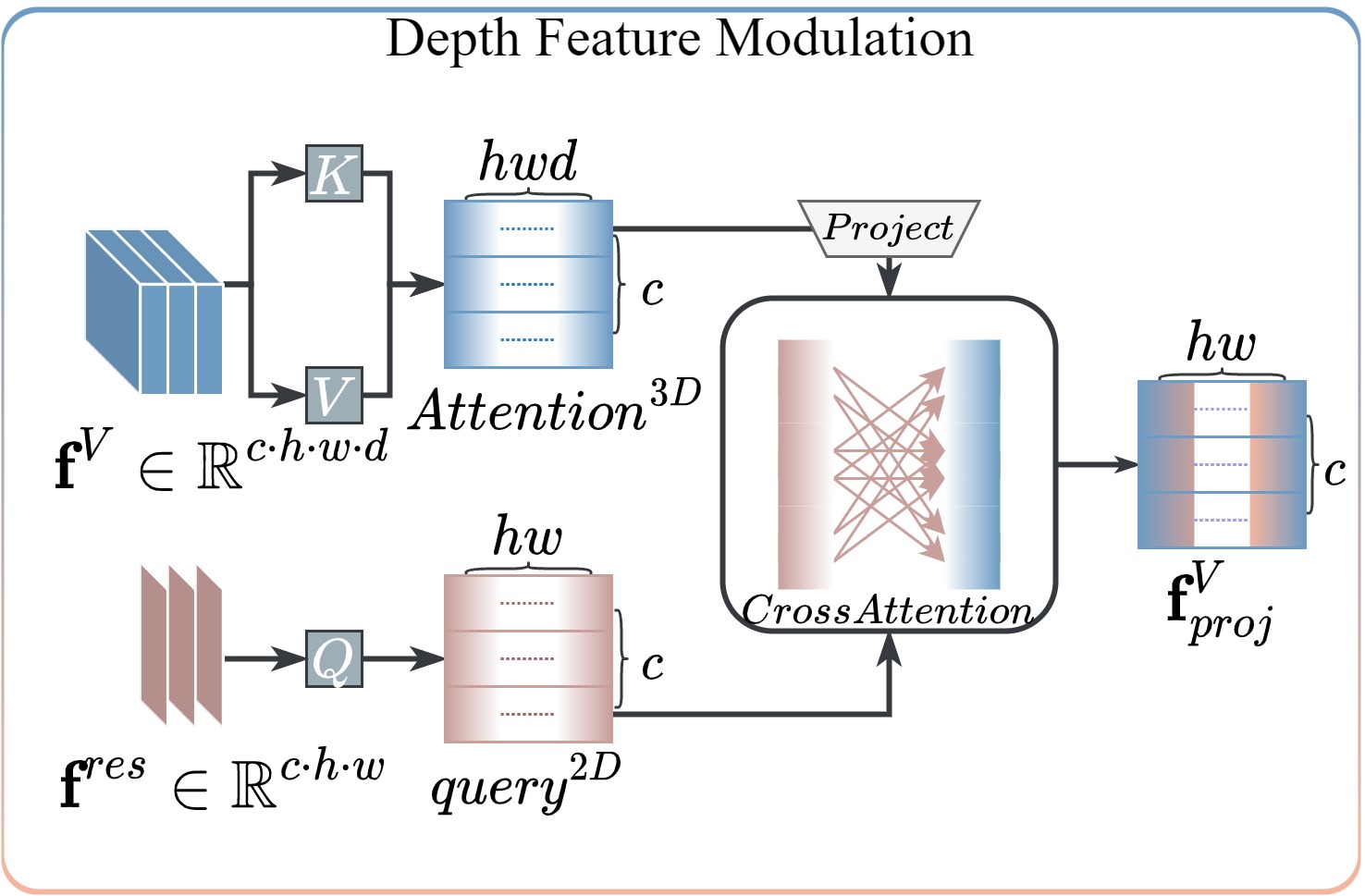}
    \caption{Depth Feature Modulation structure, we take the planar features as the query ($Q$) and the CT features as the key ($K$) and value ($V$), and achieve fusion through a cross-attention.}
    \label{fig:c}
\end{figure}

X-ray projections from any angle share the same CT, but directly fusing the entire 3D feature is redundant. Although the $A^{3d}_{enc}$ has already compressed the structural information, it still contains pose-irrelevant features. We design a screening mechanism that extracts pose-relevant components from the full CT features, as illustrated in the \cref{fig:c}.
We employ a cross-attention mechanism, taking the X-ray features as $Q$ to retrieve $K$ and $V$ from the CT features. To ensure accurate identification of the CT features corresponding to the X-ray projection, we merge the dimension-reduced \( \mathbf{f}^{V}_{proj} \) with \( \mathbf{f}^{res} \) via convolution, and feed them into the DUL reconstruction layer, where a self-supervised loss constrains the 3D feature screening process.

\subsection{Registration Network Component Design}
The core idea of image registration is to find correlated regions between the moving and fixed images and to solve for an explicit spatial transformation based on the trend of pixel changes—an ability that regressors generally lack compared to other registration models. We incorporate this idea into our design by using DCG to mine feature correlations and enhance structures, and TPM to capture pixel flow. After filtering and reorganizing the features, they are fed into the backbone for pose prediction.
\begin{figure}[htbp]
       \centering
    \includegraphics[width=0.46\textwidth]{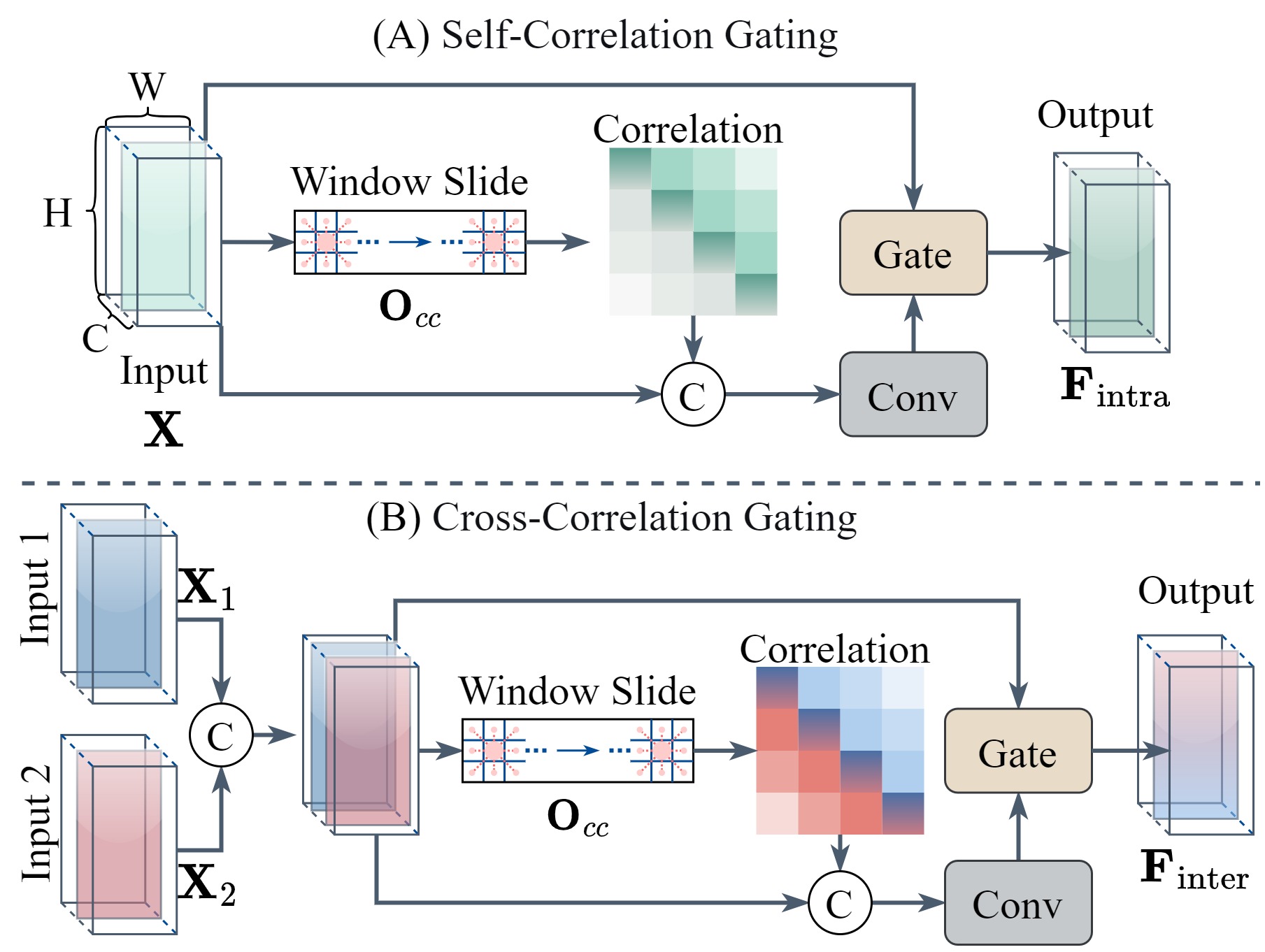}
    \caption{DCG implementation. (A) Single-input branch produces \(\mathbf{F}_{\mathrm{intra}}\). (B) Dual-input branch produces \(\mathbf{F}_{\mathrm{inter}}\). In both, \(\mathbf{O}_{cc}\) computes correlation via sliding windows over features. Concatenation with the input forms the gate \(\mathbf{G}_{\mathbf{X}}\).}
    \label{f3}
\end{figure}
\subsubsection{Dual Correlation Gate}
In recent non-rigid volumetric registration, local inner products are commonly employed to compute feature correlations and construct an explicit cost volume, thereby guiding the model to learn correspondences between moving and fixed images. DCG builds on this yet dispenses with the cost volume, serving as a gating mechanism that strengthens features from a single input and mines correlations between two inputs through correlation operators, as illustrated in \cref{f3}.

Intra‑feature gating independently amplifies the salient structures of each view, so that their subsequent subtraction in TPM can better reveal the transformation trend. For a single feature map \(\mathbf{X}\), let \(X_{p,q}^{c}\) denote the value at channel \(c\) and position \((p,q)\). For each offset \((i,j)\neq(0,0)\) within a \(3\times3\) window (\(K{=}8\) neighbours), the normalized cross‑correlation response is:
\begin{equation}
    \mathbf{O}_{cc}^{i,j}(\mathbf{X})(p,q)=\frac{(\mathbf{x}_{p,q}-\mu_{p,q})^{\top}(\mathbf{x}_{p+i,q+j}-\mu_{p+i,q+j})}
{\|\mathbf{x}_{p,q}-\mu_{p,q}\|\;\|\mathbf{x}_{p+i,q+j}-\mu_{p+i,q+j}\|},
\end{equation}
where \(\mu_{p,q}\) and \(\mu_{p+i,q+j}\) are the channel‑wise means. Stacking the responses over all \(K\) offsets yields \(\mathbf{O}_{cc}(\mathbf{X})\in\mathbb{R}^{K\times H\times W}\).
The correlation tensor is compressed and applied as a structural gate:
\begin{equation}
\mathbf{G}_{\mathbf{X}}=\mathbf{act}\big(\mathbf{Norm}(\operatorname{Conv}_{3\times3}([\mathbf{O}_{cc}(\mathbf{X}),\mathbf{X}]))\big),
\end{equation}
\begin{equation}
\mathbf{F}_{\mathrm{intra}}=\mathbf{X}+\mathbf{G}_{\mathbf{X}}\odot\mathbf{X},
\label{e7}
\end{equation}

\(\odot\) represents the Hadamard product. The gate is applied independently to \(\mathbf{X}_1\) and \(\mathbf{X}_2\), producing \(\mathbf{F}_{\mathrm{intra}}^{1},\mathbf{F}_{\mathrm{intra}}^{2}\in\mathbb{R}^{C\times H\times W}\), which are then forwarded to TPM.

Inter‑feature branch (correlation gating). The two maps are first fused into a joint representation \(\mathbf{Y}=\operatorname{Conv}_{3\times3}([\mathbf{X}_1,\mathbf{X}_2])\in\mathbb{R}^{C\times H\times W}\), and then gated by \(\mathbf{G}_{\mathbf{Y}}\), which is computed with the same gating \cref{e7}, to yield \(\mathbf{F}_{\mathrm{inter}}=\mathbf{Y}+\mathbf{G}_{\mathbf{Y}}\odot\mathbf{Y}\). This output explicitly encodes cross‑view structural agreement and is fed into regressor backbone. No dense cost volume is built.

\subsubsection{Trend Perception Module}
The TPM measures the most significant differences between two feature maps to capture deep pixel-level motion tendencies, as \cref{f2} pink part. Drawing on iterative optimization that models spatial transformations via residuals, we decompose the residual map into directional trend components in a learnable manner to assist pose regression. To obtain a global receptive field, TPM employs a multi-layer MLP. Given two DCG-enhanced feature maps \(\mathbf{F}_{\rm intra}^1, \mathbf{F}_{\rm intra}^2 \in \mathbb{R}^{C \times H \times W}\), we first compute the residual and flatten it:
\begin{equation}
    \mathbf{r} = \operatorname{Flatten}(\mathbf{F}_{\rm intra}^1 - \mathbf{F}_{\rm intra}^2) \in \mathbb{R}^{C \times HW}.
\end{equation}

Considering that translation and rotation typically affect the \(\hat{o}_x\) (horizontal) and \(\hat{o}_y\) (vertical) directions anisotropically, we explicitly swap the roles of height and width to obtain a complementary representation:
\begin{equation}
    \mathbf{s} = \operatorname{Linear}(\mathbf{Norm}(\mathbf{r})), \quad
\mathbf{s}' = \operatorname{Transpose}_{H\leftrightarrow W}(\mathbf{s}),
\end{equation}
where \(\mathbf{s}, \mathbf{s}' \in \mathbb{R}^{C \times HW}\). This representation is further processed by global response normalization (GRN) and another linear layer, yielding a transposed residual better suited for the subsequent two-branch processing:
\begin{equation}
    \mathbf{r}' = \operatorname{Linear}(\operatorname{GRN}(\mathbf{s}')) \in \mathbb{R}^{C \times HW}.
\end{equation}

To separately enhance the transformation trends along \(\hat{o}_x\) and \(\hat{o}_y\), we reshape \(\mathbf{r}'\) back to \(\mathbb{R}^{C \times H \times W}\) and construct a height-width swapped copy. The two representations are passed through independent MLPs with skip connections, producing \(\mathbf{r}_{x}\) and \(\mathbf{r}_{y}\), which are then fused via a depthwise convolution:
\begin{equation}
    \mathbf{r}_f = \mathbf{act}\bigl(\mathbf{Norm}(
\operatorname{Conv}_{3\times3}([\mathbf{r}_x, \mathbf{r}_y])
)\bigr).
\end{equation}

To clarify the direction of the trend, we use \(\mathbf{F}_{\rm intra}^1\) from the target view as a gate to adjust \(\mathbf{r}_{x}\) and \(\mathbf{r}_{y}\) and fuse the features to produce the final trend \(\mathbf{T}\):
\begin{equation}
\begin{aligned}
\mathbf{T} &= \mathbf{G}_x \odot \mathbf{r}_{_x} + \mathbf{G}_y \odot \mathbf{r}_{y} + \lambda \cdot \mathbf{r}_f + \mathbf{r}, \\
\mathbf{G}_x &= \mathbf{Norm}(\operatorname{DWConv}_{3\times1}(\mathbf{F}_{\rm intra}^1)), \\
\mathbf{G}_y &= \mathbf{Norm}(\operatorname{DWConv}_{1\times3}(\mathbf{F}_{\rm intra}^1)),
\end{aligned}
\end{equation}
where \(\lambda\) is a learnable scalar. The resulting \(\mathbf{T}\) encodes the dominant transformation trend and is fed into the subsequent pose regressor.

\subsubsection{Mamba-based Regression Backbone}
\begin{figure}
    \centering
    \includegraphics[width=0.5\textwidth]{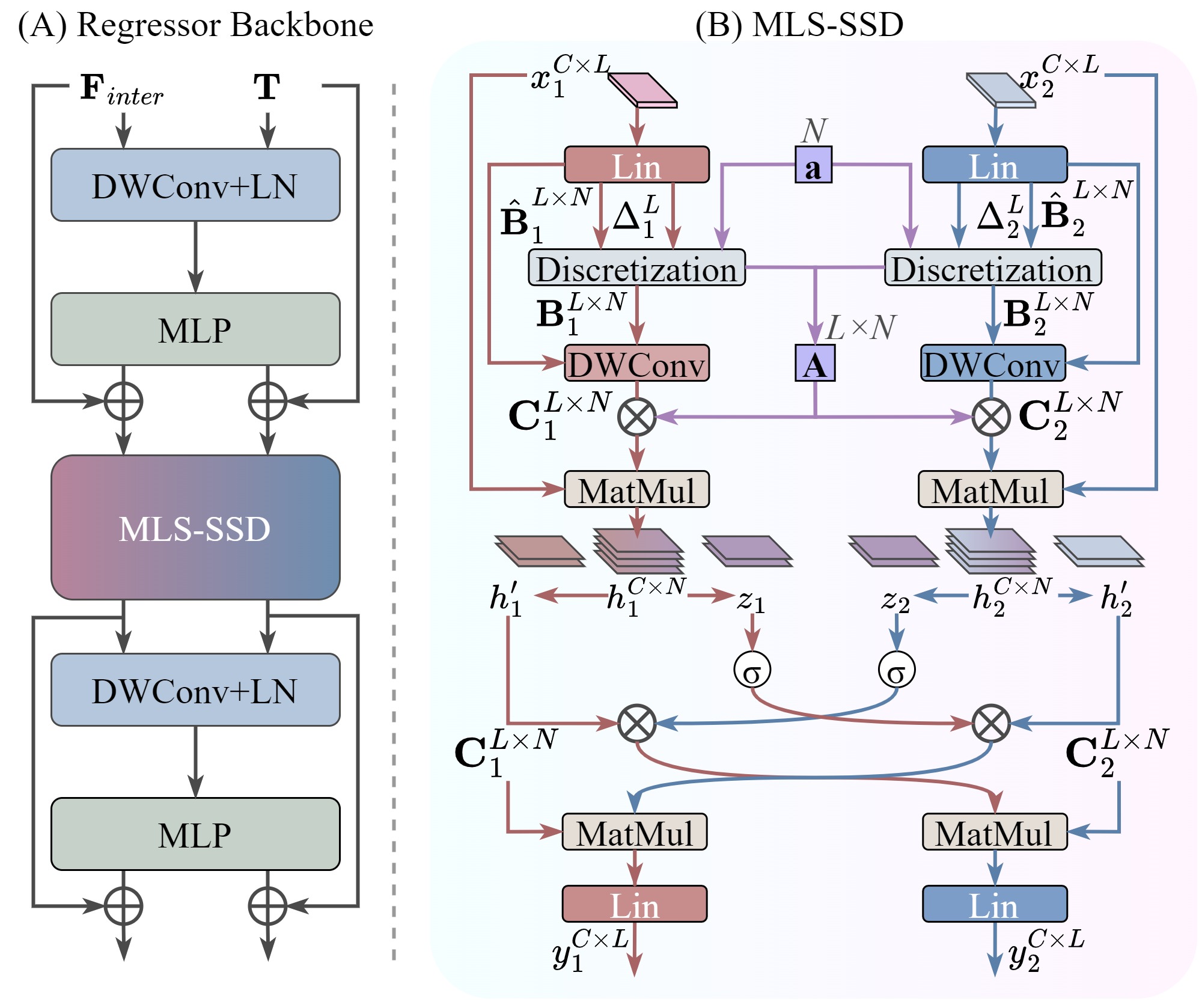}
    \caption{(A) Regression network backbone, (B) MLS-SSD implementation. Based on visual SSD by introducing mutual latent interaction between dual streams ($h_1, h_2$). Here, $C$ denotes channel number, $L$ the vector length, and $\sigma$ the gating activation function. }
    \label{f4}
\end{figure}

We propose a Mamba-based regression backbone, as shown in \cref{f4}, to perform deep interaction on two-stream inputs and predict 6DoF pose. It consists of DWConv+FFN followed by mutual latent share state space duality (MLS-SSD). The initial regression feature is obtained by concatenating \( \mathbf{T} \) and \( \mathcal{C}_{\rm inter} \) (\( \mathbb{R}^{2C \times H \times W} \)), processed by DWConv+FFN, and then split channel-wise into dual latent streams \( X_1, X_2 \in \mathbb{R}^{B \times C \times L} \).

MLS-SSD builds upon the convolution-form visual SSD~\cite{effvim, Vim} by introducing mutual latent interaction between the two streams. For each stream, input-dependent parameters are projected and refined via a shared DWConv to obtain \( \mathbf{B}_i \), \( \mathbf{C}_i \), \( \Delta_i \in \mathbb{R}^{B \times N \times L} \) (\( i=1,2 \); \( N \) is the state dimension). A shared learnable state transition vector \( \mathbf{A} \in \mathbb{R}^N \) is broadcast and combined with \( \Delta_i \) to form the selective matrix:
\begin{equation}
\mathbf{S}_i = \mathrm{softmax}\bigl( \Delta_i + \mathbf{1}_B \mathbf{A}^\top \mathbf{1}_L \bigr) \in \mathbb{R}^{B \times N \times L}, \quad i=1,2.
\end{equation}
The hidden state evolution is realized through the global convolution approximation of the selective SSM:
\begin{equation}
h_i = X_i \bigl( \mathbf{S}_i \odot \mathbf{B}_i \bigr)^\top \in \mathbb{R}^{B \times C \times N}, \quad i=1,2.
\end{equation}
Each \( h_i \) is projected and split into the main representation and gating vector \([h_i, z_i]\). A cross-gating mechanism with residual connection enables mutual interaction:
\begin{equation}
\tilde{h}_1 = h_1' \odot \mathrm{Mish}(z_2) + D \cdot h_1, \quad
\tilde{h}_2 = h_2' \odot \mathrm{Mish}(z_1) + D \cdot h_2,
\end{equation}
where \( D \in \mathbb{R} \) is a learnable scalar. The final outputs are:
\begin{equation}
y_1 = \tilde{h}_1 \mathbf{C}_2^\top, \quad y_2 = \tilde{h}_2 \mathbf{C}_1^\top.
\end{equation}
These outputs \(y_1\) and \(y_2\) are first concatenated and refined by DWConv+FFN before being reshaped back to \(\mathbb{R}^{B \times C \times H \times W}\) and passed to the next registration backbone as mutual gating signals. MLS-SSD employs bidirectional 2D scanning to preserve spatial structure while maintaining linear complexity, in contrast to the quadratic complexity of cross-attention.

\subsection{Nodes Regression and Loss Function}
LayersReg reformulates a single regression mapping as a progressive optimization process. Motivated by cascade registration paradigms~\cite{PRSCS,corrmlp}, we treat the pose output at each stage as an intermediate node whose estimate incrementally converges toward the final pose. In this way, the estimation of \(\Delta\hat{\mathcal{P}}\) is decomposed into a sequence of small, easy-to-learn steps. Each node predicts an incremental 6-DoF pose parameter \(\Delta\hat{\mathcal{P}}_i\), represented in the Lie algebra coordinates of \(\mathfrak{se}(3)\). Specifically, \(\Delta\hat{\mathcal{P}}_i\) consists of an axis-angle rotation vector (radians) and a translation vector (millimeters). The learnable weight is directly applied to this 6-DoF parameter, and the weighted increment is then mapped to \(SE(3)\) through the exponential map for group composition:
\begin{equation}
    \Delta\hat{\mathcal{P}}
    =
    \mathrm{Log}
    \left(
        \prod_{i=1}^{S}
        \mathrm{Exp}
        \left(
            \hat{\beta}_i \Delta\hat{\mathcal{P}}_i
        \right)
    \right),
    \label{eq17}
\end{equation}
where \(\mathrm{Exp}(\cdot): \mathfrak{se}(3) \rightarrow SE(3)\) maps a 6-DoF pose parameter to its rigid transformation, \(\mathrm{Log}(\cdot): SE(3)\rightarrow\mathfrak{se}(3)\) maps the composed transformation back to the pose-parameter space, and \(\prod\) denotes sequential composition on \(SE(3)\). The node weights are given by \(\hat{\boldsymbol{\beta}}=\mathrm{Softmax}(\boldsymbol{\beta})\), with learnable \(\boldsymbol{\beta}\in\mathbb{R}^{S}\). The pose output \(\Delta\hat{\mathcal{P}}_i\) at each stage is obtained by fusing dual-stream features \(y_i^1, y_i^2 \in \mathbb{R}^{C \times H \times W}\) via channel-wise gating. Specifically, the two feature maps interact through a Hadamard product, followed by pooling, normalization, and a linear regression head:
\begin{equation}
    \Delta\hat{\mathcal{P}}_i
    =
    \mathrm{Linear}
    \big(
        \mathrm{Norm}
        \big(
            \mathrm{Pool}(y_i^1 \odot y_i^2)
        \big)
    \big).
\end{equation}
This fusion reduces channel-wise redundancy and enhances pose-relevant features across stages. To mitigate intermediate error accumulation, we apply deep pose supervision to each node:
\begin{equation}
    \mathcal{L}_{pos}
    =
    \sum_{i=1}^{S}
    \left\|
        \Delta\mathcal{P}_{gt}
        -
        \Delta\hat{\mathcal{P}}^i
    \right\|_1,
\end{equation}
where \(\Delta\hat{\mathcal{P}}^i\) denotes the cumulative 6-DoF pose parameter obtained by applying \cref{eq17} to the first \(i\) weighted increments.

The 2D self‑supervision term is defined as:
\begin{equation}
    \mathcal{L}_{rec}^{2D} = \|I_{res} - I_{res}^{rec}\|_1 + 1 - \frac{2\sigma_{\nabla I_{res}\nabla I_{res}^{rec}} + \epsilon}{\sigma_{\nabla I_{res}}^2 + \sigma_{\nabla I_{res}^{rec}}^2 + \epsilon}
\end{equation}
where \(I_{res} = [I_{tar}, I_{ini}]\) is the concatenation of the target and initial views, and \(I_{res}^{rec}\) is its reconstruction from \(A^{2d}_{enc}\). Here \(\sigma_{\nabla I}^2\) is the variance of the image gradient magnitude map, \(\sigma_{\nabla I_{res}\nabla I_{res}^{rec}}\) is the covariance between the gradient maps of \(I_{res}\) and \(I_{res}^{rec}\), and \(\epsilon = 0.01\). The volumetric self‑supervision term is:
\begin{equation}
    \mathcal{L}_{rec}^{3D} = \|V_{pre} - V^{rec}\|_1,
\end{equation}
where \(V^{rec}\) is the preoperative volume reconstructed from \(A^{3d}_{enc}\). Together with the pose supervision loss \(\mathcal{L}_{pos}\), the total training loss of LayersReg combines these three terms:
\begin{equation}
    \mathcal{L} = \lambda_1\mathcal{L}_{rec}^{2D} + \lambda_2\mathcal{L}_{rec}^{3D} + \lambda_3\mathcal{L}_{pos}
    \label{eq:loss}
\end{equation}

\section{Experiments}
\subsection{Datasets and Preprocessing}
We evaluate LayersReg on seven public datasets for rigid X-ray/CT and S2V registration. All volumes are resampled to 1.0 mm isotropic spacing, cropped to \(224\times224\times192\) voxels, and intensity-normalized. A canonical initial pose is applied. Training pairs are generated by uniform 6-DoF perturbations, with relative poses as labels. Standard capture ranges are \(R\in[-15^\circ,15^\circ]\), \(T\in[-10\,\text{mm},10\,\text{mm}]\); large-misalignment ranges are \(R\in[-45^\circ,45^\circ]\), \(T\in[-20\,\text{mm},20\,\text{mm}]\).

For X-ray/CT registration, we use CTSpine1K~\cite{CTSpine1K} and VerSE~\cite{Verse} for synthetic spine X-rays, and DeepFluoro~\cite{deepflu} and Ljubljana~\cite{lju} for real clinical evaluation. For each patient CT volume, we generate DRRs from multiple distinct C-arm angles. Each DRR is paired with an X-ray acquired at a different angle, ensuring unique relative 6-DoF poses for all DRR--X-ray pairs within the same patient. We adopt a strict patient-level 8:1:1 split into training, validation, and test sets, with fully disjoint poses to prevent data leakage and anatomical overfitting. Pose labels are obtained directly from the known rigid transformations used in rendering (synthetic DRRs) or from the provided ground-truth registrations (DeepFluoro and Ljubljana). Calibration protocols differ across datasets. For CTSpine1K and VerSE, we follow the ProST~\cite{reg13} protocol to simulate the clinical C-arm environment, and for DeepFluoro and Ljubljana, we use the official clinical C-arm intrinsics and extrinsics for fair comparison. To enable rapid regressor extension, we limit its training on real X-rays to within one hour~\cite{reg14}.

For S2V, we use RESECT~\cite{RESECT} (MRI-US), a cardiac dataset~\cite{xingzang} (MRI-CT, 43 volumes), and CAMUS~\cite{CAMUS} (real US). In RESECT, expert-annotated landmarks provide cross-modal transforms with TRE $\le$ 1 mm. Slices are extracted along probe directions with recorded 6-DoF poses. For dense supervision, 10,000 slices per patient are randomly sampled. All S2V datasets adopt an 8:1:1 patient-level split, with augmentations to mimic clinical conditions. CAMUS preprocessing follows prior works~\cite{CUreg,FVR,EUreg}.

\subsection{Implementation Details}
The LayersReg framework is implemented in PyTorch~\cite{pytorch} and trained/tested on two NVIDIA A800 GPUs with 160 GB memory. LayersReg has 34.49 M parameters, 58.39 G FLOPs, and an inference time of 101.8 ms. For the S2V task, the single-frame alignment time is 0.21 s, while for X-ray/CT alignment it is 0.14 s, fully meeting real-time requirements. The initial learning rate is set to \(1 \times 10^{-3}\), and the optimizer is AdamW + GC~\cite{GC} (weight decay 0.05) with cosine annealing scheduler decaying to \(1 \times 10^{-6}\). X-ray/CT registration is trained for 100 epochs on a million-scale dataset, while the S2V task requires 50 epochs. A batch size of 128 is employed, with early stopping after 30 epochs of non-decreasing loss. The primary loss \(\mathcal{L}_{\rm pos}\) is augmented with scaled auxiliary terms to ensure balanced optimization: the 2D reconstruction loss \(\mathcal{L}^{2d}_{\rm rec}\) is scaled by 0.1 and \(\mathcal{L}^{3d}_{\rm rec}\) by 0.5. These terms are fused via learnable uncertainty parameters \(\lambda_1, \lambda_2, \lambda_3\), yielding the composite loss \cref{eq:loss}. Ablation studies confirm this approach outperforms grid-search hyperparameter optimization. For 6-DoF pose estimation, we use MAE(R) (mean absolute Euler angles, °) and Dist(T) (Euclidean distance, mm). SSIM, NCC, and RMSE evaluate structural similarity between the warped preoperative volume—generated via DRR projection for X-ray/CT or reslicing for S2V using the predicted pose—and the ground-truth aligned intraoperative image. These metrics are computed in the registered common coordinate system after alignment, measuring structural quality rather than raw inter-modal appearance similarity.

\begin{table}[htpb]
    \centering
    \small
    \setlength{\tabcolsep}{3pt}
    \caption{Performance of LayersReg in Different Complex Challenges}
    \begin{tabular}{lcccccc}
      \toprule
      \multirow{2}{*}{Task} &
      \multirow{2}{*}{Offset} &
      \multirow{2}{*}{Type} &
      \multicolumn{2}{c}{MAE (R)} &
      \multicolumn{2}{c}{Dist (T)} \\
      \cmidrule(lr){4-5} \cmidrule(lr){6-7}
      & & & Mean & Std & Mean & Std \\
      \midrule
      \multirow{3}{*}{Xray/CT}
        & large & None & 0.64 & 0.27 & 1.41 & 1.68 \\
        & standard & Occluded1 & 0.58 & 0.31 & 1.55 & 1.66 \\
        & large & Occluded2 & 0.72 & 0.44 & 2.51 & 1.70 \\
      \midrule
      \multirow{3}{*}{S2V}
        & standard & US-MRI & 0.38 & 0.25 & 0.84 & 0.56 \\
        & standard & CT-MRI & 0.31 & 0.13 & 0.66 & 0.48 \\
        & large & FLAIR-T1 & 0.55 & 0.20 & 1.57 & 1.05 \\
      \bottomrule
    \end{tabular}
    \label{t1}
\end{table}
\subsection{Adaptability to Complex Challenges}
\begin{figure}[htbp]
    \centering
    \includegraphics[width=0.45\textwidth]{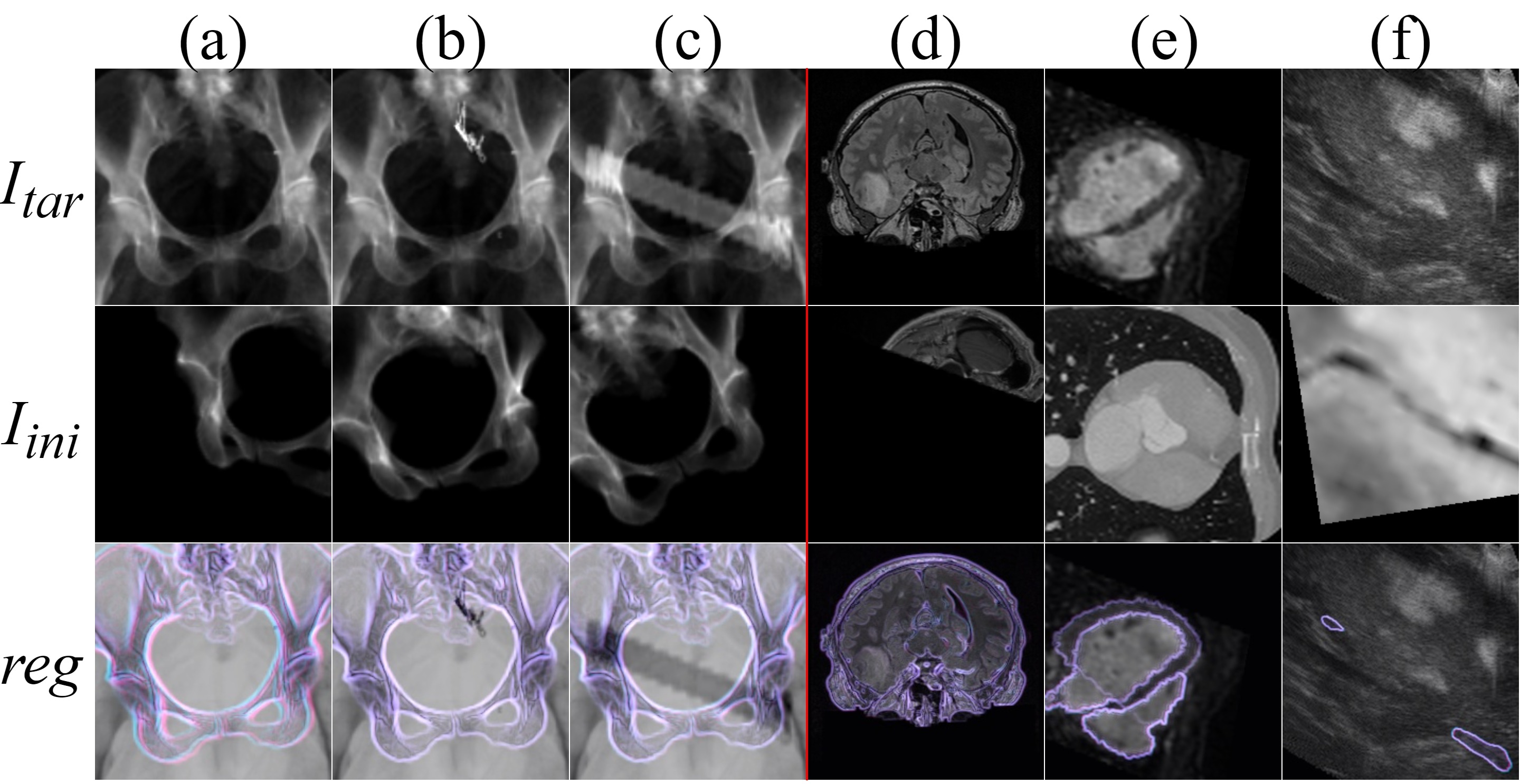}
    \caption{Registration performance of LayersReg across various application scenarios: (a) large displacement, (b) occlusion by pedicle screw, (c) occlusion type 2, (d) FLAIR-T1 brain MRI, (e) MRI-CT ventricle, (f) intraoperative craniotomy ultrasound to preoperative MRI registration.}
    \label{f5}
\end{figure}

To evaluate the adaptability of LayersReg, we selected challenging scenarios prevalent in real-world environments for testing, including X-ray images with real surgical instruments occlusions, multi-modal slice alignment, and large offset errors. The results are presented in \cref{t1} and \cref{f5}. We overlay the boundaries (red and cyan) of the ground-truth and prediction results on $I_{tar}$, purple is the aligned contour. 

The results demonstrate that LayersReg is adaptable across 3D/2D registration tasks involving diverse modalities, achieving high-precision alignment in terms of predicted poses and registration similarity metrics. To further assess the capture range, we extend the standard-range setting to large-offset scenarios and evaluate the success rate under gradually enlarged perturbation intervals. Large misalignment usually requires back-projection-based registration, whereas LayersReg performs alignment without additional projection reconstruction, benefiting from depth-aware feature compensation and progressive internal refinement. Since landmark-based mTRE is not available for all datasets, failure is defined by pose errors: a case is regarded as failed if the final rotation error exceeds \(5^\circ\) or the translation error exceeds \(10\) mm. Under this criterion, LayersReg maintains reliable registration up to \([-60^\circ,60^\circ]\) rotation and \([-50,50]\) mm translation for X-ray/CT, and up to \([-50^\circ,50^\circ]\) rotation and \([-30,30]\) mm translation for S2V.

\subsection{Ablation Experiments}
\subsubsection{Regression strategy study}
We conduct ablation studies on regression strategies, comparing our node-based method with three baselines: direct regression, averaging across stages, and uncertainty-based selection. Results in Table~\ref{t2} show that our approach achieves the lowest errors (MAE (R): 0.37°, Dist (T): 1.71 mm) and highest stability, as it effectively simulates iterative refinement. The alternatives yield higher errors or less stable predictions, confirming the suitability of node-based regression for LayersReg.

\begin{table}[htbp]
\centering
\label{tab:ablation}
\centering
\caption{Regression Strategies}
\label{t2}
\begin{tabular}{ lcccc }
\toprule
\multirow{2}{*}{Models} & \multicolumn{2}{c}{MAE (R)} & \multicolumn{2}{c}{Dist (T)}\\
\cmidrule(lr){2-3} \cmidrule(lr){4-5}
& mean & std & mean & std \\
\midrule
Nodes      & \textbf{0.37} & \textbf{0.25} & \textbf{1.71} & \textbf{1.33} \\
Mean      & 0.43 & 0.27 & 2.06 & 1.44 \\
Uncertain & 0.42 & 0.28 & 1.97 & 1.35 \\
Direct    & 0.41 & 0.28 & 1.92 & 1.53 \\
\bottomrule
\end{tabular}
\end{table}

\begin{table}[htbp]
\centering
\caption{Ablation on progressive module integration.}
\label{tab:ablation_incremental}
\begin{tabular}{ lcccc }
\toprule
\multirow{2}{*}{Models} & \multicolumn{2}{c}{MAE (R)} & \multicolumn{2}{c}{Dist (T)}\\
\cmidrule(lr){2-3} \cmidrule(lr){4-5}
& mean & std & mean & std \\
\midrule
Baseline  & 2.79 & 1.89 & 5.81 & 3.28 \\
+ $A^{3d}_{enc}$     & 1.55 & 1.29 & 3.50 & 2.84 \\
+ DCG             & 1.05 & 0.98 & 2.46 & 2.24 \\
+ TPM                & 0.71 & 0.83 & 1.94 & 1.87 \\
+ MLS-SSD      & \textbf{0.46} & \textbf{0.32} & \textbf{1.47} & \textbf{1.42} \\
\bottomrule
\end{tabular}
\end{table}

\subsubsection{Progressive Module Ablation}
We assess the contribution of each component through a progressive ablation study. The baseline is a standard 3D-2D CNN regressor without any specialized modules (MAE(R)=2.79°, Dist(T)=5.81 mm). As shown in \cref{tab:ablation_incremental}, successively adding each module leads to consistent performance gains. Introducing the 3D encoder ($A^{3d}_{\rm enc}$) provides essential depth information. Adding DCG strengthens feature correspondence and correlation. Integrating TPM effectively captures transformation trends from residuals. Finally, MLS-SSD enables global modeling with linear complexity, attaining the full LayersReg performance with MAE(R)=0.46° and Dist(T)=1.47 mm. These results confirm that every proposed module contributes meaningfully and progressively to the overall regression accuracy.

\subsubsection{Loss functions study}
To validate the optimality of our proposed loss combination, we conducted ablation studies by removing individual components: the 2D/3D self-supervised losses and the contrastive geodesic loss \(\mathcal{L}_{geo}\)~\cite{reg2}. Results are shown in Table~\ref{t3}.

\begin{table}[htbp]
    \centering
    \caption{Ablation experiments with loss functions}
    \begin{tabular}{@{}cccccccc@{}}
      \toprule
      \multicolumn{4}{c}{Loss Functions} & \multicolumn{2}{c}{MAE (R)} & \multicolumn{2}{c}{Dist (T)} \\
      \cmidrule(lr){1-4} \cmidrule(lr){5-6} \cmidrule(lr){7-8}
      $\mathcal{L}^{3d}_{rec}$ & $\mathcal{L}^{2d}_{rec}$ & $\mathcal{L}_{pos}$ & $\mathcal{L}_{geo}$ & mean & std & mean & std \\
      \midrule
      & $\checkmark$ & $\checkmark$ & & 0.46 & 0.32 & 2.13 & 1.71 \\
      $\checkmark$ & & $\checkmark$ & & 0.41 & \textbf{0.27} & 1.89 & 1.48 \\
      $\checkmark$ & $\checkmark$ & & $\checkmark$ & 0.43 & 0.30 & 1.92 & 1.46 \\
      $\checkmark$ & $\checkmark$ & $\checkmark$ & & \textbf{0.39} & 0.28 & \textbf{1.78} & \textbf{1.34} \\
      $\checkmark$ & $\checkmark$ & $\checkmark$ &$\checkmark$ & 0.40 & 0.31& 1.79 & 1.47 \\
      \bottomrule
    \end{tabular}
    \label{t3}
\end{table}

The dual self-supervised losses combined with \(\mathcal{L}_{pos}\) achieve optimal performance, confirming \cref{eq:loss} as the most suitable for LayersReg. Ablating the 3D reconstruction loss causes the largest degradation (MAE(R) +0.09, Dist(T) +0.42), highlighting the critical role of depth cues for large-offset robustness in S2V tasks. Removing the 2D loss reduces robustness to X-ray noise and occlusions. Adding \(L_{geo}\) does not further improve accuracy, suggesting that the direct pose loss is sufficient under our 6-DoF parameterization. Overall, our learnable weighted loss best balances accuracy and adaptability.

\subsection{Compare with the SOTA Methods}
In this section, we compare our method with state-of-the-art 3D/2D registration approaches under two experimental settings. The first setting performs alignment on real ultrasound frames and X-ray images, while the second leverages synthetic X-ray projections and ultrasound slices. Considering that spine intervention involving multimodal data represents a practical clinical scenario, yet real datasets remain scarce, evaluation on simulated data serves as a viable alternative. For fairness, all methods were evaluated using identical patient-level splits, capture ranges, and metrics within each experimental group. \cref{t4} reports task-level pooled results, where all held-out test cases from the corresponding datasets are aggregated within each task category.

\begin{figure}
    \centering
    \includegraphics[width=0.9\textwidth]{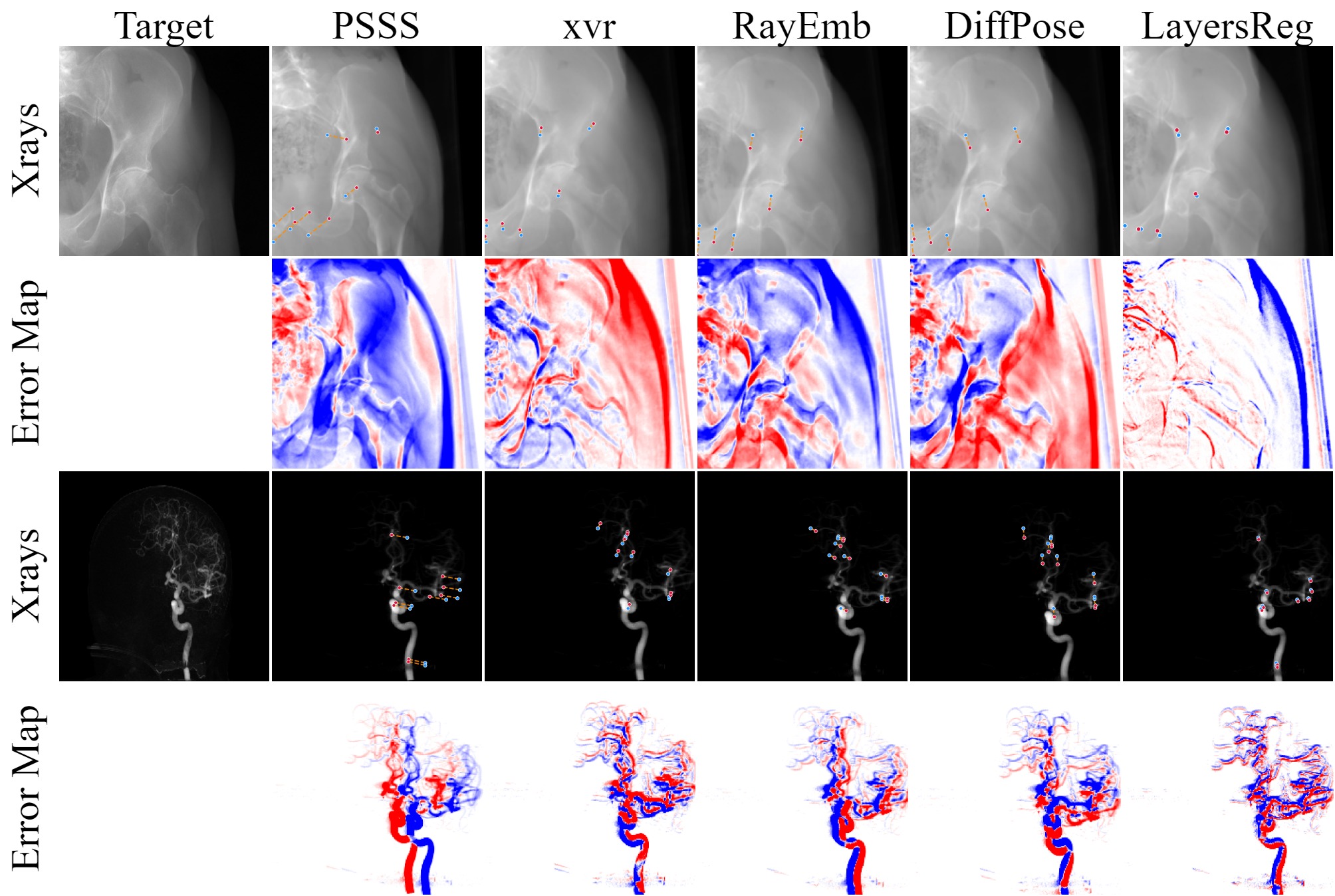}
    \caption{Compared with SOTA methods on surgery X-rays. Orange lines indicate landmark offset; the 2nd and 4th rows show error maps (vs. GT).}
    \label{f8}
\end{figure}
\subsubsection{Evaluation on Surgical Scenarios}
In this section, we evaluate on intraoperative X-ray images under a strict 1-hour training budget for all methods, simulating the time-critical nature of intraoperative deployment. Methods tailored for these datasets are adopted as baselines, and we report both mTRE \cite{mTRE} and registration success rate (mTRE \(\le 10\) mm). The results are shown in \cref{f8} and \cref{tab:final}.
\begin{table}[htbp]
    \centering
    \caption{Intraoperative SOTA Comparison}
    \label{tab:final}
    \footnotesize
    \setlength{\tabcolsep}{2.5pt}
    \begin{tabular}{lcccccc}
        \toprule
        \multirow{2}{*}{Methods} & \multirow{2}{*}{Type} & \multicolumn{2}{c}{DeepFluoro \cite{deepflu}} & \multicolumn{2}{c}{Ljubljana \cite{lju}} \\
        \cmidrule(lr){3-4} \cmidrule(lr){5-6}
        & & mTRE (mm)& SR (\%) & mTRE (mm) & SR (\%) \\
        \midrule
        PSSS \cite{reg14}    & Regressor          & 25.27 & 23 & 27.93 & 19 \\
        xvr \cite{xvr}       & Regressor          & 5.25  & 91 & 7.16  & 68 \\
        DiffPose \cite{reg2} & Regressor + opt    & 8.25  & 62 & 9.87  & 46 \\
        RayEmb \cite{RayEmb} & Landmark + opt     & 6.49  & 74 & 8.94  & 59 \\
        \midrule
        \textbf{LayersReg}   & Regressor          & \textbf{4.59} & \textbf{95} & \textbf{5.63} & \textbf{91} \\
        \bottomrule
    \end{tabular}
\end{table}

As shown in \cref{tab:final}, LayersReg outperforms all baseline methods. Specifically, xvr and DiffPose adopt online self-supervised training, yet they still lag behind. On Ljubljana, LayersReg reduces mTRE by 1.53 mm over xvr. Moreover, unlike RayEmb, which requires costly landmark annotations, LayersReg achieves accurate alignment without manual labeling and obtains the best mTRE and success rate.

\subsubsection{Extensive Evaluation on Multiple Scenarios}
\begin{figure*}
    \centering
    \includegraphics[width=0.9\textwidth]{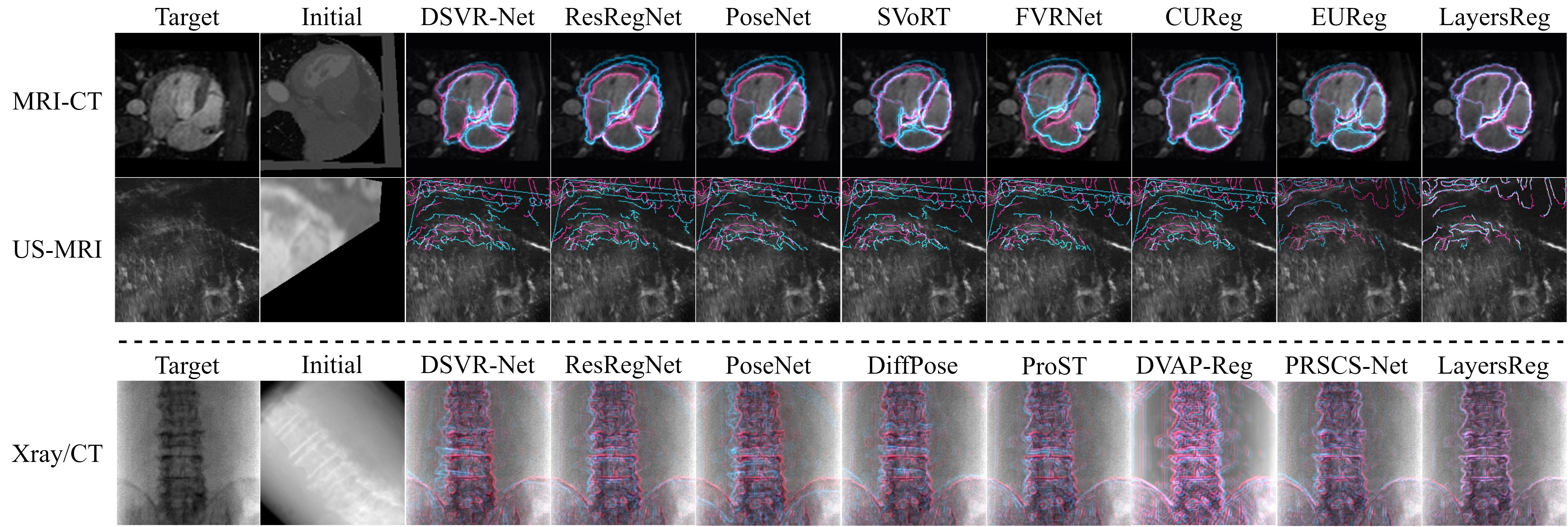}
    \caption{Visualization of the comparison with the SOTA 3D/2D registration network. Red, blue, and purple contours represent the ground truth, prediction, and overlap.}
    \label{f7}
\end{figure*}

In this set of experiments, we evaluate on multi-modality (US-MRI) registration tasks, including US-MRI and CT-MRI, as well as X-ray to CT registration for the spine. The baselines cover standard pose regression networks such as Res-RegNet and PoseNet, orthogonal-view registration networks such as DVAP and PRSCS, and the recent differentiable registration method DiffPose. The results are presented in \cref{f7} and \cref{t4}.

\begin{table}[]
    \centering
    \caption{Compare with the SOTA 3D/2D registration methods in different tasks}
    \label{t4}
    \footnotesize
    \setlength{\tabcolsep}{1.8pt}
    \begin{tabular}{l|cccccccc}
        \toprule
        \multirow{2}{*}{Type} &
        \multirow{2}{*}{Methods} & 
        \multicolumn{2}{c}{MAE(R)} & \multicolumn{2}{c}{Dist(T)} &  \multirow{2}{*}{NCC} & \multirow{2}{*}{RMSE}  & \multirow{2}{*}{SSIM} \\
        \cmidrule(lr){3-4} \cmidrule(lr){5-6}
        & & mean & std & mean & std & & & \\
        \midrule
        \textbf{S2V}
        & DSVR-net \cite{regsota} & 3.06 & 3.05 & 8.66 & 7.34 & 0.83 & 0.12 & 0.82 \\
        & Res-RegNet \cite{reg4} & 3.69 & 3.66 & 9.96 & 7.57 & 0.82 & 0.13 & 0.81 \\
        & SVoRT \cite{SVort} & 3.11 & 2.97 & 9.68 & 7.16 & 0.84 & 0.12 & 0.82  \\
        & PoseNet \cite{reg12} & 4.52 & 3.44 & 13.66 & 7.59 & 0.79 & 0.15 & 0.78 \\
        & FVRNet \cite{FVR} & 7.65 & 4.49 & 6.67 & 5.54 & 0.78 & 0.17 & 0.74 \\
        & CUReg\cite{CUreg} & 3.19 & 3.48 & 5.96 & 5.47 & 0.83 & 0.11 & 0.84 \\
        & EUReg \cite{EUreg} & 1.39 & 1.34 & 2.78 & 1.77 & 0.92 & 0.09 & 0.88\\
        & \textbf{LayersReg} & \textbf{0.73} & \textbf{0.46} & \textbf{1.55} & \textbf{0.91} & \textbf{0.97} & \textbf{0.05} & \textbf{0.93} \\
        \midrule
        \textbf{Xray/CT}
        & DSVR-net \cite{regsota} & 2.75 & 2.36 & 4.98 & 3.18 & 0.88 & 0.09 & 0.52 \\
        & Res-RegNet \cite{reg4} & 1.31 & 0.59 & 7.23 & 3.44 & 0.87 & 0.10 & 0.50 \\
        & PoseNet \cite{reg12} & 1.75 & 0.85 & 8.14 & 4.05 & 0.81 & 0.11 & 0.41 \\
        & ProST \cite{reg13} & 1.71 & 1.94 & 3.19 & 2.27  & 0.89 & 0.09 & 0.61\\
        & DiffPose\cite{reg2} & 1.56 & 1.86 & 2.99 & 2.75 & 0.90 & 0.08 & 0.63  \\
        & DVAP-Reg\cite{DVAP-Reg} & 1.78 & 0.96 & 1.75 & 1.58 & 0.91 & 0.07 & 0.65 \\
        & PRSCS-Net \cite{PRSCS} & 1.35 & 1.08 & 1.93 & 1.16 & 0.93 & 0.06 & 0.71 \\
        & \textbf{LayersReg} & \textbf{0.68} & \textbf{0.57} & \textbf{1.41} & \textbf{0.98} & \textbf{0.96} & \textbf{0.05} & \textbf{0.79} \\
        \bottomrule
    \end{tabular}
\end{table}
The proposed LayersReg achieves better performance across both S2V and X-ray/CT tasks as a pure regressor. In S2V, it outperforms EUReg by reducing MAE(R) from 1.39° to 0.73° and Dist(T) from 2.78 mm to 1.55 mm, while improving NCC from 0.92 to 0.97 and SSIM from 0.88 to 0.93. In X-ray/CT, LayersReg surpasses PRSCS-Net, lowering MAE(R) from 1.35° to 0.68° and Dist(T) from 1.93 mm to 1.41 mm, with NCC rising from 0.93 to 0.96 and SSIM from 0.71 to 0.79. These consistent gains highlight LayersReg's superior accuracy and robust structural alignment in heterogeneous multimodal 3D/2D registration.

\begin{figure}
    \centering
    \includegraphics[width=0.5\textwidth]{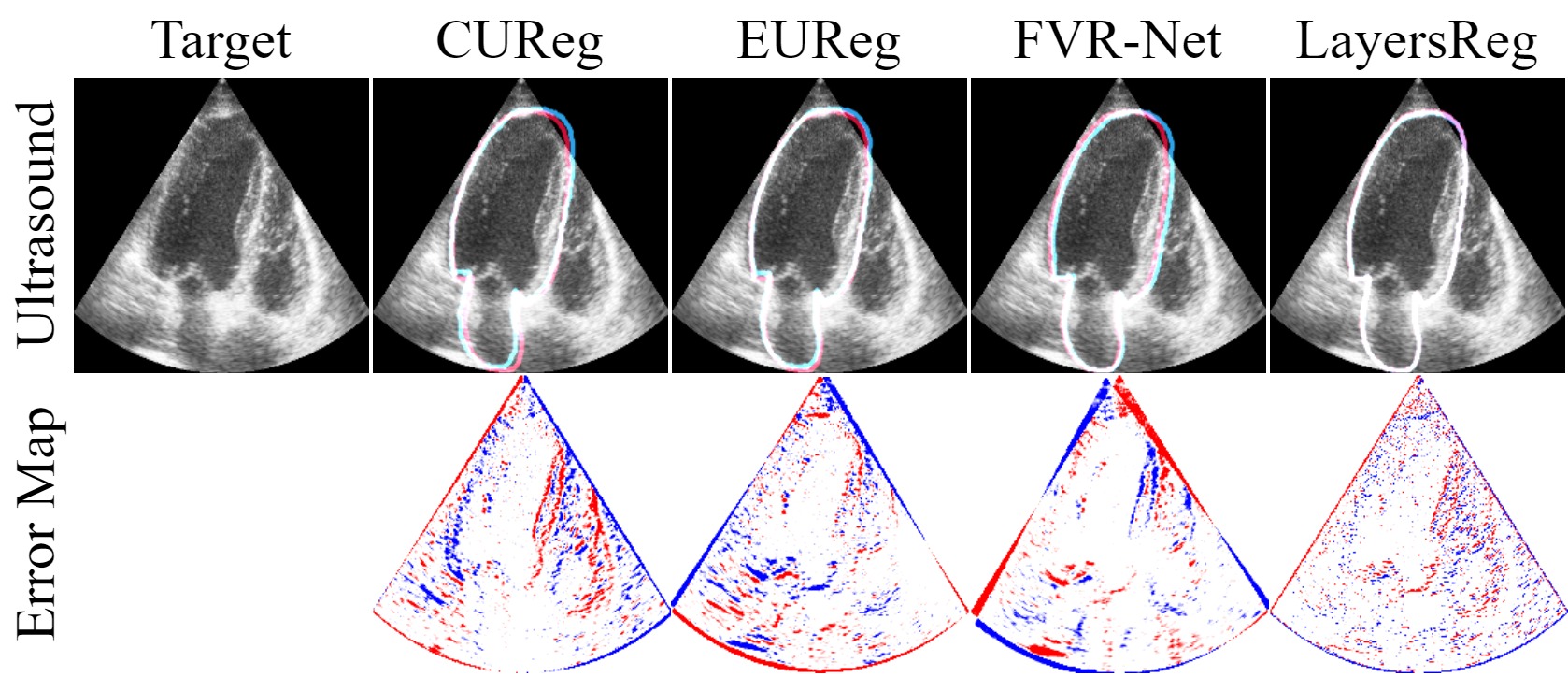}
    \caption{
Slice localization comparison with ground truth on the CAMUS dataset, red is ground truth, blue is prediction.}
    \label{f9}
\end{figure}

Notably, we present an additional registration case on the CAMUS dataset with real US frame (\cref{f9}) to evaluate the accuracy of the S2V task. LayersReg achieves the most complete ventricular contour alignment with the smallest errors.

\section{Conclusion}
We propose LayersReg, a novel 3D/2D registration framework. It reinterprets regression as an internal progressive optimization process. By modeling feature correlations and transformation trends, LayersReg captures the regularities of accurate pose correction. It outperforms existing methods on both X-ray/CT and S2V registration tasks. Extensive experiments show that, even in complex intraoperative scenarios, LayersReg does not require physicians to manually delineate RoIs. The framework can be integrated with advanced differentiable rendering and further refined through test-time optimization to achieve pixel-level accuracy. Moreover, it enables rapid adaptation to new patients, with only the regressor fine-tuned while the autoencoders remain frozen, and one hour of training sufficing for deployment. The current model handles rigid transformations only. In the future, we plan to extend its optimization-driven mechanism to deformable 3D/2D registration for soft tissues, a promising direction for further exploration.

\section*{Acknowledgments}
This work was supported by the State Key Laboratory of Digital Medical Engineering – Jiangsu Province Matching Funding (Grant No. 4207032303) and the Jiangsu Youth Natural Science Foundation (Grant No. BK20241277).

\bibliographystyle{unsrt}  
\bibliography{references}

\end{document}